\documentclass[letterpaper, 10 pt, conference]{ieeeconf}
\usepackage{kotex}

\IEEEoverridecommandlockouts
\usepackage{amsmath, amssymb}
\usepackage{graphicx}
\usepackage{multirow}
\usepackage{cite}
\usepackage{tabularx}
\usepackage{makecell}

\title{\LARGE \bf
MSC: Multi-Stage Caster-aware Control Framework\\
with Uncertainty-Aware NMPC and RL-based Execution Layer
}

\author{Jinho Park$^{*}$, Jeongtae Lee$^{*}$,  Jeonghyeok Lim$^{*}$, Keun Ha Choi$^{\dag}$ and Kyung-Soo Kim$^{\dag}$ 
\thanks{*These authors contributed equally to this work.}
\thanks{\dag Corresponding authors.}
\thanks{The authors are with the Department of Mechanical Engineering, KAIST, Daejeon, Republic of Korea
        {\tt\small \{ppp109403, us03316, henricus0973, choiha99, kyungsookim\}@kaist.ac.kr}}%
}

\begin{document}
\maketitle
\thispagestyle{empty}
\pagestyle{empty}



\begin{abstract}


Freely swiveling casters are widely adopted in industrial AMRs/SDVs for their mechanical simplicity and load capacity. However, caster misalignment during low-speed or high-curvature maneuvers induces strong contact nonlinearities, widening the planning-execution gap between commanded $(v, \omega)$ and realized motion. Existing caster-aware strategies are highly sensitive to state-estimation reliability, causing unstable execution in uncertain transition regimes. To address this, we present Multi-Stage Caster-aware Control (MSC), a planning-execution integrated hierarchical framework featuring two core mechanisms. First, an uncertainty-aware NMPC (CU-MPC) dynamically schedules caster-aware costs and input regularization using predictive uncertainty to suppress overly aggressive maneuvers. Second, a reinforcement learning-based execution layer estimates and predicts caster states under partial observability, transforming high-level velocity commands into physically executable wheel speeds adapted to current contact conditions. Validations in Isaac Sim and real robots demonstrate that MSC offers a clear comparative advantage over existing decoupled compensation approaches by significantly improving caster-state estimation accuracy, minimizing misalignment, and enhancing tracking robustness in caster-sensitive maneuvers.
\end{abstract}

\section{Introduction}
\label{sec:introduction}


Industrial AMR/SDV navigation performance is not determined solely by the quality of the planner. In confined logistics environments, a robot must not only generate collision-free trajectories that satisfy actuator constraints, but also ensure that the underlying drivetrain can stably realize the planned motion in practice. Although many platforms employ freely swiveling caster wheels, contact interactions in caster-equipped nonholonomic drivetrains can substantially increase the required drive-wheel torque and power; consequently, the executability of planned velocity commands may degrade, particularly during low-speed turning and in-place rotations~\cite{Staicu2009NonlinearDynamics,Chung2003ICRA_CasterDynamics}. In particular, caster alignment mismatch induces reaction torques and additional rolling resistance, which can lead to trajectory deviation and poor velocity tracking, and in severe cases, stall events~\cite{Arrizabalaga2021CasterAwareMPC,Sprigle2019ScrubTorque}.

\begin{figure}[t]
  \centering
\includegraphics[width=\linewidth]{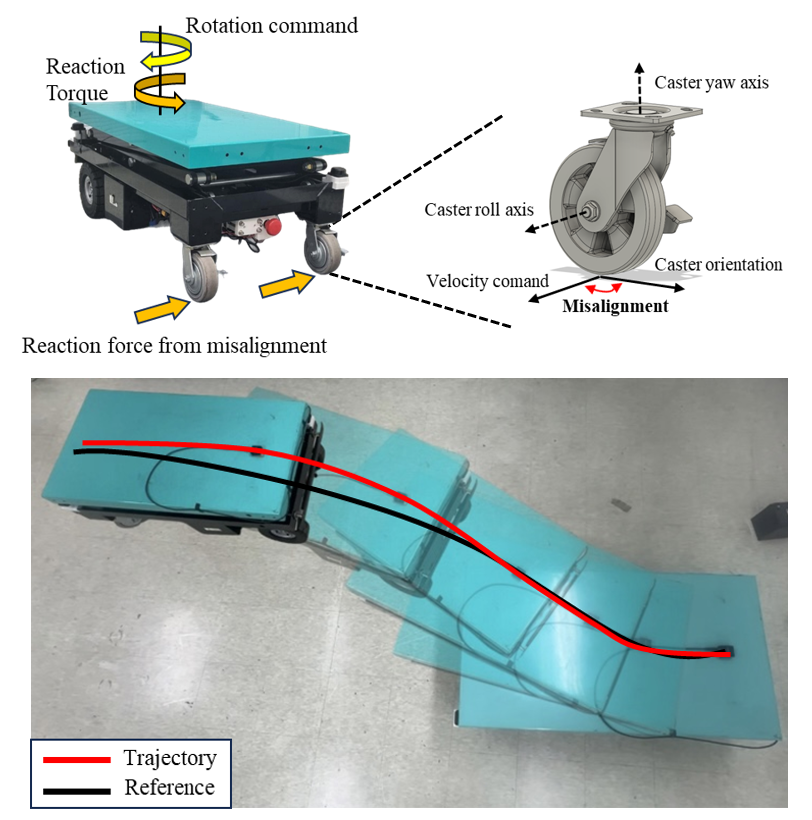}
  \caption{\noindent\textbf{Caster misalignment and its navigation impact.}
A mismatch between the commanded motion direction and the caster yaw angle direction generates a lateral scrub contact force and a resistive caster-induced reaction torque, contributing to a planning--execution gap between planned $(v,\omega)$ and physically realizable motions.}
  \label{fig:problem_overview}
  \vspace{-1.0em} %
\end{figure}


Prior studies can be broadly categorized into two lines:
(i) \textit{decoupled compensation}, which treats caster-induced reaction torques as disturbances at the low-level control layer and compensates for them within velocity/torque tracking controllers~\cite{WangZhai2020DOB,Nishizawa2025LyapunovCertifiedTailWheel,Schroeder2019EnhancedMotionControl}, and
(ii) \textit{caster-aware planning}, which explicitly incorporates caster effects into cost terms and constraints at the motion-planning or NMPC level~\cite{Arrizabalaga2021CasterAwareMPC}.
Approach (i) often post-processes or retrospectively adjusts the planner-generated velocity commands, which can introduce a mismatch between the idealized dynamics assumed during planning and the corrective actions taken during execution.
Approach (ii) is highly sensitive to the accuracy of caster-state estimation (e.g., alignment angle and rolling condition); as estimation errors grow, the resulting plans may become overly conservative or may fail during execution.

In real-world settings, variations in payload and friction, as well as micro-slip phenomena, can significantly increase the uncertainty of caster-state estimation.
If such uncertainty is not explicitly reflected in high-level planning, the planner may produce commands that are not physically executable, and relying solely on downstream compensation at the execution layer is insufficient to simultaneously guarantee both robustness to estimation uncertainty and physical realizability.
Therefore, \emph{uncertainty-aware planning} and \emph{execution-consistent realization} should be considered jointly across the hierarchy.

To this end, we propose MSC (\textit{Multi-Stage Caster-aware Control}), a hierarchical framework that integrates caster-state estimation/prediction and its associated uncertainty into high-level planning, while ensuring that the planner's $(v,\omega)$ commands are realized in a manner consistent with the current contact conditions at the low-level execution layer.
This integration mitigates the planning--execution gap.





\subsection*{Contributions}
The main contributions of this paper are as follows:
\begin{enumerate}
  \item \textbf{MSC: Planning--execution integrated framework} \\
  We propose MSC, a hierarchical framework that jointly addresses caster-aware planning and execution under contact nonlinearities.

  \item \textbf{Caster Uncertainty-aware NMPC (CU-MPC)} \\
  We introduce horizon-wise uncertainty-aware scheduling of the caster-aware cost and input regularization to reduce misalignment in uncertain transition regimes.

  \item \textbf{RL-based execution layer} \\
  We develop an RL-based execution layer that estimates/predicts caster states with uncertainty and transforms high-level $(v,\omega)$ into executable wheel-speed commands, improving tracking in low-speed maneuvers.
\end{enumerate}

\vspace*{0.4em}



\section{Related Works}
\label{sec:related_work}
\subsection{Disturbance Rejection and Velocity/Torque Command Compensation}
\label{subsec:rw_decoupled}

Previous studies have mitigated motion degradation caused by wheel--ground interactions and passive-wheel dynamics through disturbance compensation, robust tracking, and command shaping.
Wang and Zhai combined a disturbance observer with adaptive control to compensate for skidding, slipping, and lumped disturbances in wheeled mobile robots~\cite{WangZhai2020DOB}.
Nishizawa \textit{et al.} developed a differential-flatness-based tracking controller with adaptive backstepping and Lyapunov stability guarantees for a tail-wheel-equipped robot~\cite{Nishizawa2025LyapunovCertifiedTailWheel}.
Schr{"o}der \textit{et al.} proposed a model-based path filter that reshapes motion commands using estimated caster orientations and rolling states to reduce current peaks and caster locking~\cite{Schroeder2019EnhancedMotionControl}.

Although effective for tracking and disturbance rejection, these methods compensate for caster effects after the desired trajectory or command has been determined.
They therefore cannot jointly optimize caster reorientation, trajectory objectives, collision avoidance, and actuator constraints during planning, motivating planning-level caster-aware approaches.


\subsection{Caster-aware Planning at the High-level Planning Stage}
\label{subsec:rw_planning}


Arrizabalaga et al. proposed a caster-wheel-aware MPC-based motion planner that incorporates a bore-torque-related cost and uses an observer to estimate caster angle and rolling speed ~\cite{Arrizabalaga2021CasterAwareMPC,Arrizabalaga2020ThesisCWAWLP}. This formulation biases the optimized trajectory away from maneuvers requiring substantial caster reorientation and was evaluated against decoupled bore-torque reduction and caster-agnostic approaches. However, the planner relies on deterministic observer estimates and does not explicitly propagate predictive uncertainty over the MPC horizon. CU-MPC extends this formulation by predicting short-horizon caster-state distributions and adapting the caster-aware cost and input regularization according to the estimated uncertainty.


\subsection{Learning-Based State Estimation for Control}
\label{subsec:rw_learning}


Learning-based methods infer unobserved states or environment properties from
observation histories under partial observability.
Kumar et al. estimated latent environment information for policy adaptation
~\cite{kumar2021rma}, while Ji et al. jointly trained a state estimator and a
control policy~\cite{ji2022concurrent}.
Gu et al. recovered internal dynamic representations from noisy or incomplete
observations using a denoising world model~\cite{gu2024advancing}.

These approaches mainly use inferred information to improve learned locomotion
policies.
In contrast, our method estimates passive caster states, predicts their
short-horizon mean and variance, and provides the resulting uncertainty to
CU-MPC.

\vspace*{0.4em}
\begin{figure*}[t]
  \centering
  \includegraphics[width=\textwidth]{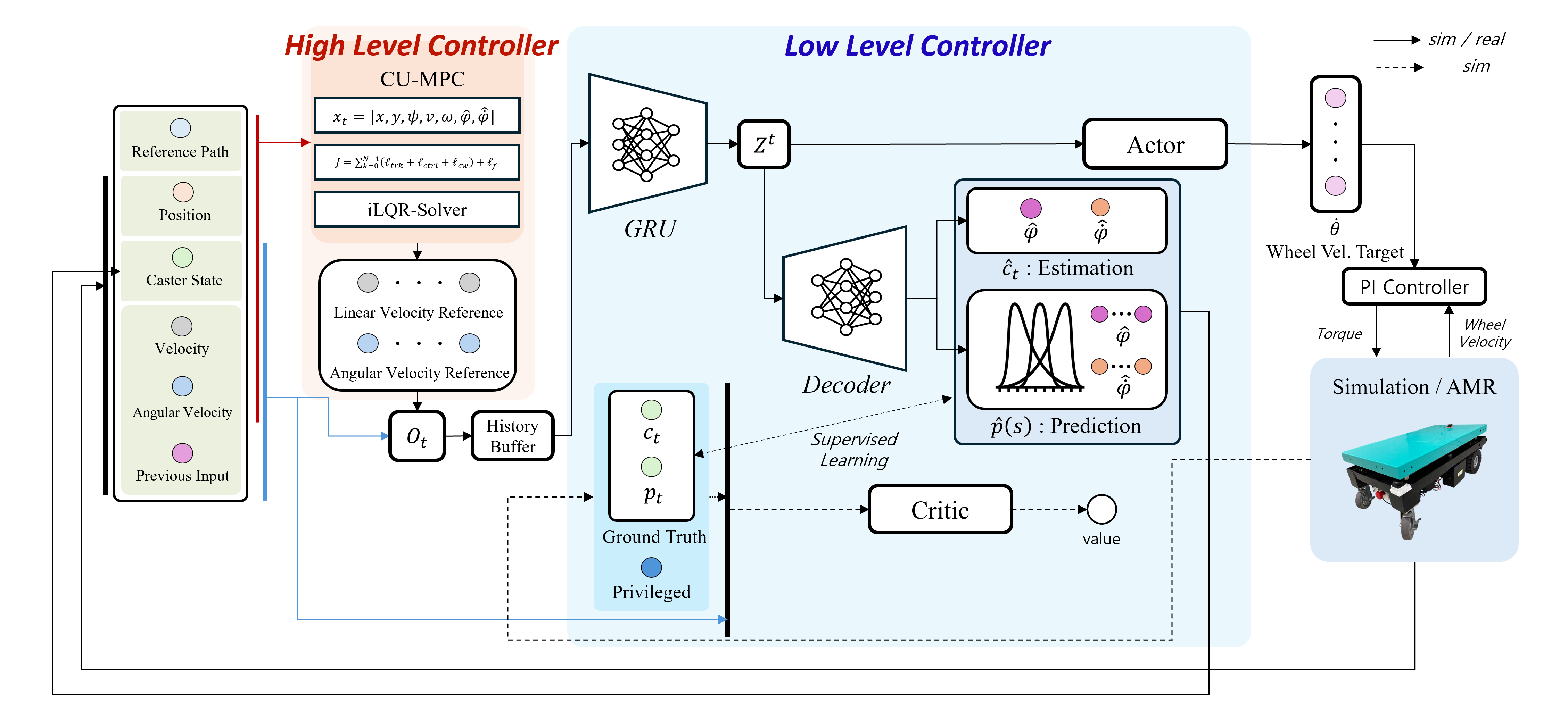}
  \caption{\textbf{Overall framework of MSC.}
The caster-uncertainty-aware CU-MPC generates local velocity references.
An RL-based Execution Layer, trained with privileged information in simulation,
estimates/predicts caster states with uncertainty and transforms the high-level
$(v,\omega)$ commands into dynamically feasible wheel commands.
This hierarchical integration mitigates the planning--execution gap
arising from caster-induced contact nonlinearities.}
  \label{fig:overall_framework}
  \vspace{-1.0em} %
\end{figure*}



\section{Methods}
\label{sec:method}



\subsection{Overview of Control Framework}
\label{subsec:method_framework}
Passive internal joints such as freely swiveling casters cannot be directly regulated by control inputs.
Variations in their states can abruptly alter both the required actuation torque and the feasible control set of the overall system, particularly in singularity-like regimes or during caster re-alignment.
In such conditions, uncertainty in real-time state estimation can directly affect execution stability.

To address this challenge, we propose an integrated hierarchical framework that couples caster-state estimation and short-horizon prediction in the low-level controller with uncertainty-aware model-based planning at the high level, as illustrated in Fig.~\ref{fig:overall_framework}.
The framework consists of (i) a \textbf{high-level controller} and (ii) a \textbf{low-level controller}.
The high-level controller builds upon the caster-aware NMPC formulation in prior work and retains the extended state (including caster states) and the basic form of the caster-aware cost term~\cite{Arrizabalaga2021CasterAwareMPC}.
In contrast to kinematics-based observers, the proposed framework employs a learning module in the low-level controller to provide caster-state \textit{estimation/prediction}, and further introduces an \textit{uncertainty-aware} extension that conservatizes planning over the prediction horizon using predictive uncertainty.

At runtime step $t$, the high-level controller generates an optimized velocity command over the NMPC horizon $N$ based on the state estimate $\hat{\mathbf{x}}_t$ and the reference sequence $\mathbf{r}_{t:t+N}$:
$\mathbf{u}^{\mathrm{ref}}_{t|t}\!=\![v^{\mathrm{ref}}_{t|t},\omega^{\mathrm{ref}}_{t|t}]^\top$.
The low-level controller infers a latent state $\mathbf{z}_t$ from the observation history $\mathbf{o}_t^{(H)}$ and outputs:
(1) the caster-state estimate $\hat{\mathbf{x}}^{\mathrm{cw}}_t$,
(2) $H_p$-step caster-state predictions with confidence indicators $\{(\hat{\mathbf{x}}^{\mathrm{cw}}_{t+k|t},\,\sigma_{t+k|t})\}_{k=1}^{H_p}$, and
(3) an executable command $\mathbf{u}^{\mathrm{exec}}_t$ corresponding to the high-level velocity command $\mathbf{u}^{\mathrm{ref}}_{t|t}$.
That is, the low-level controller modifies the high-level command so that $\mathbf{u}^{\mathrm{ref}}_{t|t}$ can be realized consistently under caster-induced contact nonlinearities, and applies the resulting command to the drivetrain.


\vspace{0.3em}
\noindent
\textbf{Online execution loop:}
\begin{enumerate}
  \item Sensing and state estimation: acquire $\hat{\mathbf{x}}_t$ and $\mathbf{o}_t^{(H)}$
  \item High-level control (CU-NMPC) optimization: compute $\mathbf{u}^{\mathrm{ref}}_{t|t}$
  \item Caster state estimation and prediction: obtain $\hat{\mathbf{x}}^{\mathrm{cw}}_t$ and
$\{(\hat{\mathbf{x}}^{\mathrm{cw}}_{t+k|t},\,\sigma_{t+k|t})\}_{k=1}^{H_p}$.
  \item Low-level command mapping: $\mathbf{u}^{\mathrm{ref}}_{t|t}\mapsto \mathbf{u}^{\mathrm{exec}}_t$
  \item Execution: apply $\mathbf{u}^{\mathrm{exec}}_t$ and repeat
\end{enumerate}

To prevent algebraic loops and optimization delays, the CU-MPC at step $t$ uses the previous cycle's buffered predictions and uncertainty $\{(\hat{x}_{t+k|t-1}^{cw}, \sigma_{t+k|t-1})\}$.


\subsection{High-Level Controller: Caster Uncertainty-Aware MPC}
\label{subsec:method_mpc}

\subsubsection{Baseline caster-aware MPC formulation}
\label{subsubsec:mpc_baseline}
The high-level controller is based on a trajectory-tracking NMPC that includes a caster-aware cost term~\cite{Arrizabalaga2021CasterAwareMPC}.
At time $t$, given the state estimate $\hat{\mathbf{x}}_t$ and the reference sequence $\mathbf{r}_{t:t+N}$, the controller optimizes the decision variables
$\mathbf{U}_t \triangleq \{\mathbf{u}_{t|t},\dots,\mathbf{u}_{t+N-1|t}\}$.
The predicted state trajectory is initialized with $\mathbf{x}_{t|t}=\hat{\mathbf{x}}_t$ and evolves according to
\begin{align}
\mathbf{x}_{t+k+1|t} = f\!\left(\mathbf{x}_{t+k|t},\mathbf{u}_{t+k|t}\right),\qquad k=0,\dots,N-1.
\label{eq:mpc_dyn}
\end{align}
The objective is defined as
\begin{equation}
\begin{aligned}
\min_{\mathbf{U}_t}\;\;
J_t \triangleq
\sum_{k=0}^{N-1}\!\Big(
&\ell_{\mathrm{trk}}(\mathbf{x}_{t+k|t};\mathbf{r}_{t+k})
+\ell_{\mathrm{ctrl}}(\mathbf{u}_{t+k|t}) \\
&+\ell_{\mathrm{cw}}(\mathbf{x}_{t+k|t})
\Big)
+\ell_f(\mathbf{x}_{t+N|t}).
\end{aligned}
\label{eq:mpc_baseline_obj}
\end{equation}
subject to \eqref{eq:mpc_dyn} and state/input/collision-avoidance constraints.
Here, $\ell_{\mathrm{cw}}(\cdot)$ penalizes caster misalignment by measuring the discrepancy between the \emph{ideal caster alignment} required to execute the target velocity command $(v^{\mathrm{ref}},\omega^{\mathrm{ref}})$ without slip and the current (estimated) caster state (e.g., caster yaw angle).
As the mismatch between the required rolling direction and the actual (estimated) caster direction increases, contact reaction forces and resistive torques can grow; accordingly, $\ell_{\mathrm{cw}}(\cdot)$ is designed to suppress such misalignment.


\vspace{0.25em}
\noindent
This formulation is effective under the assumption that the caster-state estimate is sufficiently reliable.
However, in \emph{singularity-prone} regimes---such as standstill-to-motion transitions, abrupt turns, and forward--reverse switching---limited observability can increase the uncertainty of caster-state estimation/prediction.
In such cases, the misalignment penalty defined relative to the ideal alignment state becomes less trustworthy, which may degrade tracking performance and induce numerical sensitivity (e.g., inconsistent shifts between conservative and aggressive plans).


\subsubsection{Horizon-wise Uncertainty-Aware Weight Scheduling}
\label{subsubsec:mpc_prediction_integration}
We retain the baseline caster-aware MPC structure and incorporate, into the horizon-wise cost weighting, the future caster-state predictions and uncertainty information provided by the predictor in the low-level execution layer.
At time $t$, the predictor outputs the following quantities up to $H_p$ steps ahead:
\begin{align}
\Big\{
&\hat{\varphi}_{L,t+h|t},\hat{\varphi}_{R,t+h|t},
\widehat{\dot{\gamma}}_{L,t+h|t},\widehat{\dot{\gamma}}_{R,t+h|t}, \nonumber\\
&\sigma^2_{\varphi L,t+h|t},\sigma^2_{\varphi R,t+h|t},
\sigma^2_{\dot{\gamma}L,t+h|t},\sigma^2_{\dot{\gamma}R,t+h|t}
\Big\}_{h=1}^{H_p}.
\label{eq:pred_outputs}
\end{align}
Here, $\hat{\varphi}_{(\cdot)}$ and $\widehat{\dot{\gamma}}_{(\cdot)}$ denote the predicted caster swivel angle and rolling angular velocity, respectively.

We summarize these outputs into a single risk scalar. For $h=1,\ldots,H_p$, we define
\begin{align}
\rho_{t+h|t}
\triangleq
\left(\sigma^2_{\varphi (L,R),t+h|t}\right)
+\eta\left(\sigma^2_{\dot{\gamma}(L,R),t+h|t}\right)
\label{eq:risk_scalar}
\end{align}
where $\eta>0$ is a weighting factor that balances the relative contributions of the swivel angle ($\varphi$) and rolling-rate ($\dot{\gamma}$) uncertainties.
The risk signal $\rho_{t+h|t}$ typically increases in \emph{singularity-prone} regimes such as start-up, abrupt maneuvers, and forward--reverse switching, and can be interpreted as an indicator of reduced reliability of the caster-aware term.



\vspace{0.25em}
\noindent\textbf{Caster-aware weight scheduling.}
When predictive uncertainty is high, the caster-aware term becomes less reliable.
To prevent it from overly dominating the objective, we schedule its weight over the prediction horizon:
\begin{align}
K &\triangleq \min(N, H_p), \label{eq:K_def}\\
w_{\mathrm{cw}}(t+k)
&=
w_{\mathrm{cw}}^{0}\, g\!\left(\rho_{t+k|t}\right),
\qquad k=1,\dots,K,
\label{eq:mpc_uncertainty_weighting_a}\\
g(\rho)
&=
g_{\min}+(1-g_{\min})\exp(-\beta\rho),
\label{eq:mpc_uncertainty_weighting_b}
\end{align}
where $g_{\min}\in(0,1]$ is a lower bound and $\beta>0$ is a decay rate.
Accordingly, the caster-aware term is weighted as
\begin{align}
\ell_{\mathrm{cw}}(\mathbf{x}_{t+k|t})
\;\Rightarrow\;
w_{\mathrm{cw}}(t+k)\,\ell_{\mathrm{cw}}(\mathbf{x}_{t+k|t}).
\label{eq:cw_weighted}
\end{align}

\vspace{0.25em}
\noindent\textbf{Input regularization.}
To suppress torque-sensitive maneuvers (e.g., aggressive acceleration/deceleration and abrupt turns) in the same regimes, we strengthen the input penalty horizon-wise:
\begin{align}
w_{\mathrm{ctrl}}(t+k)
=
w_{\mathrm{ctrl}}^{0}\left(1+\alpha\,\rho_{t+k|t}\right),
\qquad k=1,\dots,K,
\label{eq:mpc_ctrl_weighting}
\end{align}
where $\alpha>0$ controls the degree of conservatism.
For example, if $\ell_{\mathrm{ctrl}}(\mathbf{u})=\mathbf{u}^\top \mathbf{R}\mathbf{u}$, then \eqref{eq:mpc_ctrl_weighting} is equivalent to increasing the effective scale of $\mathbf{R}$ in uncertainty-prone segments.
Such soft regularization can be applied continuously in iLQR/NMPC and encourages more executable and consistent plans by attenuating excessive input variations under high uncertainty.

\subsection{Low-Level Controller: RL-based Execution Layer}
\label{subsec:rl_execution_layer}
\subsubsection{Network Architecture}
To infer the hidden dynamics of the caster wheel, the proposed model adopts a recurrent encoder--decoder architecture based on temporal input sequences.
\paragraph{Observation.}
The single-step observation vector is defined as
\begin{align}
\mathbf{o}_t = 
[
\widehat{\dot{\varphi}}_{L,t},\, 
\widehat{\dot{\varphi}}_{R,t},\, 
v_t,\, 
\omega_t,\, 
u_{L,t-1},\, 
u_{R,t-1},\, 
v^{\mathrm{ref}}_t,\, 
\omega^{\mathrm{ref}}_t
]^\top \in \mathbb{R}^8.
\end{align}
To incorporate temporal context, the past $H$ samples are subsampled with stride 2 to construct the discrete sequence
\begin{align}
\mathbf{o}_t^{(H)}
=
\left\{
\mathbf{o}_{t-2(H-1)},\ldots,\mathbf{o}_t
\right\}.
\end{align}

\paragraph{Temporal Encoder.}
A GRU-based encoder $f_{\mathrm{enc}}$ compresses the observation history into a latent representation:
\begin{align}
\mathbf{z}_t = f_{\mathrm{enc}}(\mathbf{o}_t^{(H)})
\end{align}
The latent vector $\mathbf{z}_t$ summarizes recent dynamic information and captures the nonlinear behavior of the caster mechanism.
It serves as the shared input to both the actor and critic networks.

\paragraph{State Estimation Head.}
The estimation module $f_{\mathrm{est}}$ reconstructs the current caster state from the latent representation:
\begin{align}
\hat{\mathbf{x}}^{\mathrm{cw}}_t
= f_{\mathrm{est}}(\mathbf{z}_t).
\label{eq:est_head}
\end{align}
In this work, the privileged caster state is defined as
\begin{align}
\hat{\mathbf{x}}^{\mathrm{cw}}_t
=
[
\hat{\varphi}_{L,t},\,
\hat{\varphi}_{R,t},\,
\widehat{\dot{\varphi}}_{L,t},\,
\widehat{\dot{\varphi}}_{R,t}
]^\top
\in \mathbb{R}^4,
\label{eq:cw_state_def}
\end{align}
where  $\hat{\varphi}_{(\cdot),t}$ and $\widehat{\dot{\varphi}}_{(\cdot),t}$ denote
the estimated swivel angle and swivel angular velocity of the left and right caster wheels, respectively.
This state is not directly observable from $\mathbf{o}_t$
and is only available in simulation as privileged information
for supervised training.
\paragraph{Probabilistic Prediction Head.}
The prediction module models the future evolution of caster swivel angles and swivel angular velocities over a horizon of $H_p$ steps.
For each prediction step $k = 1, \ldots, H_p$, the conditional distribution is modeled as
\begin{align}
p\!\left(
\mathbf{x}^{\mathrm{cw}}_{t+k}
\mid
\mathbf{z}_t
\right)
=
\mathcal{N}
\big(
\boldsymbol{\mu}_{t+k},
\mathrm{diag}(\boldsymbol{\sigma}^2_{t+k})
\big),
\label{eq:pred_gauss}
\end{align}
where $p(\cdot)$ denotes the probability density function.
The mean $\boldsymbol{\mu}_{t+k}$ and variance $\boldsymbol{\sigma}^2_{t+k}$ are directly predicted by the decoder.
The variance term quantifies predictive uncertainty in future caster trajectories,
which is subsequently transmitted to the upper-level MPC to assess the risk of entering singularity-like regimes.
\begin{figure*}[t]
  \centering
  \includegraphics[width=\textwidth]{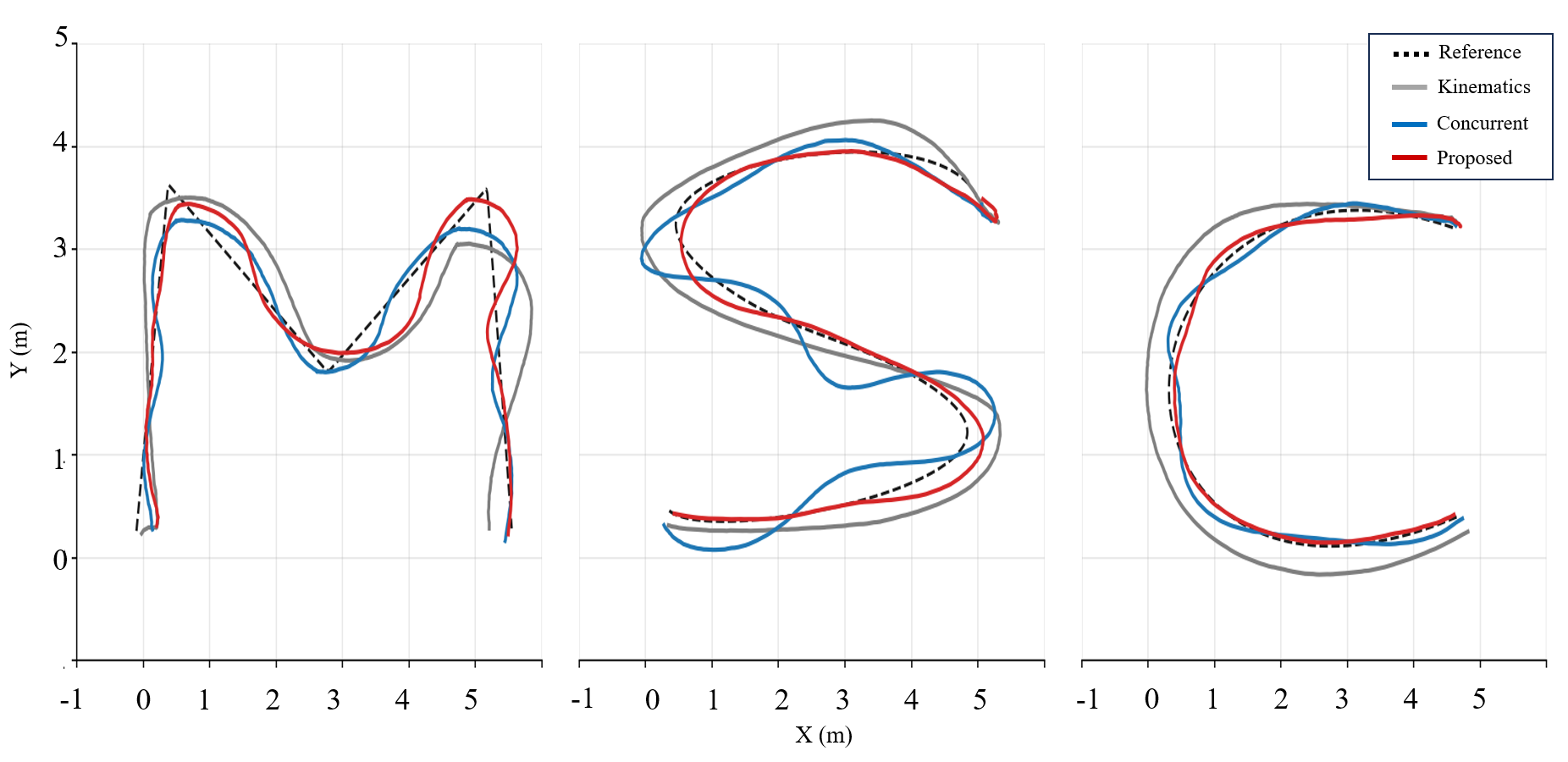}
  \caption{\textbf{Trajectory tracking comparison against baselines.}
Proposed Method improves tracking smoothness and accuracy over B1/B2, particularly under caster-induced contact nonlinearities.}
  \label{fig:tracking_results}
\end{figure*}

\paragraph{Execution Policy Head.}

The same latent representation $\mathbf{z}_t$ is used as the input to a PPO-based actor--critic policy.
The policy network is implemented as a multi-layer perceptron (MLP) with hidden dimensions $[512,256,128]$.
The actor outputs the execution command $\mathbf{u}^{\mathrm{exec}}_t$ applied to the drive motors,
while the critic estimates the state value function $V(\mathbf{z}_t)$:
\begin{align}
\mathbf{u}^{\mathrm{exec}}_t = \pi_{\mathrm{map}}(\mathbf{z}_t),
\qquad
V_t = V(\mathbf{z}_t).
\label{eq:exec_head}
\end{align}
In this work, the execution command is defined as $\mathbf{u}^{\mathrm{exec}}_t = [{\dot{\theta}_{L,t}}, {\dot{\theta}_{R,t}}]^\top$, which specifies the desired angular velocities of the left and right drive motors. On the real platform, each motor tracks ${\dot{\theta}_{(\cdot),t}}$ using a low-level PI velocity controller.
The overall architecture jointly infers the current caster state and its future probabilistic evolution from partial observations.
The estimated state provides an interpretable reconstruction of latent contact dynamics,
while the predicted mean and variance characterize both expected trajectory evolution and its uncertainty.
This uncertainty-aware representation enables the upper-level MPC
to perform risk-sensitive cost scheduling and adaptive constraint modulation
near singularity-prone regimes.

\subsubsection{Training Objective}
\label{subsubsec:training_objective}
The execution layer jointly optimizes state estimation,
probabilistic future prediction, and execution policy
within a unified objective:
\begin{align}
\mathcal{L}
=
\underbrace{
\|\hat{\mathbf{x}}^{\mathrm{cw}}_t - \mathbf{x}^{\mathrm{cw}}_t\|_2^2
+
\lambda_r \|\mathbf{z}_t\|_1
}_{\mathcal{L}_{\mathrm{est}}}
+
\lambda_{\pi} \mathcal{L}_{\pi}
+
\nonumber\\
\lambda_{V} \mathcal{L}_{V}
+
\lambda_{\mathrm{pred}} \mathcal{L}_{\mathrm{NLL}}.
\label{eq:prediction_loss}
\end{align}
Here, $\mathcal{L}_{\mathrm{est}}$ denotes the estimation loss
that reconstructs the privileged caster state
$\mathbf{x}^{\mathrm{cw}}_t$ from observation history.
The term $\lambda_r \|\mathbf{z}_t\|_1$ regularizes the latent representation
to encourage compact and informative encoding.
$\mathcal{L}_{\pi}$ and $\mathcal{L}_{V}$ correspond to
the standard PPO actor and critic losses, respectively.

\vspace{0.25em}
\noindent\textbf{Probabilistic Prediction Loss.}
The future caster prediction loss $\mathcal{L}_{\mathrm{NLL}}$
is defined as the Gaussian negative log-likelihood over an $H_p$-step horizon:
\begin{align}
\mathcal{L}_{\mathrm{NLL}}
&=
\frac{1}{H_p}
\sum_{k=1}^{H_p}
\frac{1}{2}
\Big[
(\mathbf{x}^{\mathrm{cw}}_{t+k}
-
\boldsymbol{\mu}_{t+k})^\top
\boldsymbol{\Sigma}_{t+k}^{-1}
(\mathbf{x}^{\mathrm{cw}}_{t+k}
-
\boldsymbol{\mu}_{t+k})
\nonumber\\
&\qquad
+
\log
\left|
\boldsymbol{\Sigma}_{t+k}
\right|
\Big].
\label{eq:nll_loss}
\end{align}
The covariance is parameterized as a diagonal matrix:
\begin{align}
\boldsymbol{\Sigma}_{t+k}
=
\mathrm{diag}(\boldsymbol{\sigma}^2_{t+k})
\end{align}

Here, $\boldsymbol{\mu}_{t+k}$ and $\boldsymbol{\sigma}^2_{t+k}$
are directly predicted by the decoder,
and $\mathbf{x}^{\mathrm{cw}}_{t+k}$ denotes the ground-truth
future privileged caster state obtained from simulation.

By jointly optimizing the mean and variance,
the model learns both the expected trajectory evolution
and its predictive uncertainty.
The estimated uncertainty is subsequently used by the upper-level MPC
to perform risk-aware cost scheduling,
enabling adaptive constraint relaxation or penalization
near singularity-prone regimes.

\begin{table}[t]
\centering
\caption{Key parameters used in simulation, learning, and MPC.}
\label{tab:exp_params}
\setlength{\tabcolsep}{4pt}
\renewcommand{\arraystretch}{1.08}
\begin{tabularx}{\columnwidth}{@{}l c >{\raggedright\arraybackslash}X@{}}
\hline
\textbf{Parameter} & \textbf{Symbol} & \textbf{Value} \\
\hline
\multicolumn{3}{@{}l}{\textbf{MPC (CA-MPC / CU-MPC)}} \\
NMPC horizon & $N$ & 63 \\
Sampling time & $\Delta t$ & 0.05 \\
State weight (tracking) & $\mathbf{Q}$ & 1.0 \\
Input weight (base) & $\mathbf{R}$ & 0.1 \\
Caster-aware weight & $w_{\mathrm{cw}}^{0}$ & 1.0 \\
Control weight scaling & $w_{\mathrm{ctrl}}^{0}$ & 1.0 \\
Uncertainty weighting params & $g_{\min},\beta,\alpha$ & 0.15, 1.0, 1.4 \\
Input limits & \makecell{$\mathbf{u}_{\min},\mathbf{u}_{\max}$} & (-1.2, 1.2) \\
Solver settings (iLQR) & \makecell{max iter, tol,\\ LM $\lambda$, line-search} & (20, 0.01, 1e-4, 8)  \\
\hline
\multicolumn{3}{@{}l}{\textbf{Learning (Execution module)}} \\
History length & $H$ & $16$ \\
Latent state dimension & $\dim(z_t)$ & $48$ \\
Prediction horizon & $H_p$ & 8 \\
Loss weights & \makecell{$(\lambda_{\mathrm{r}},\lambda_{\mathrm{pred}},$\\$\lambda_{\pi},\lambda_{V})$} & $(0.002, 1.0, $  $ 5.0, 5.0)$ \\
\hline
\multicolumn{3}{@{}l}{\textbf{Domain Randomization (Isaac Sim)}} \\
Wheel friction coefficient & $\mu_{\mathrm{wheel}}$ & $[0.2,\,0.9]$ \\
Caster swivel friction coefficient  & $\mu_{\mathrm{cw}}$ & $[0,\,0.1]$ \\
Caster swivel damping  & $d_{\mathrm{cw}}$ & $[0,\,0.1]$ \\
Drive joint damping ($K_d$)  & $K_d$ & $[10,\,40]$ \\
\hline
\end{tabularx}
  \vspace{-1.0em} 
\end{table}


\section{Experiments and Results}
\label{sec:experiments}

In this section, we validate the proposed MSC framework through simulation in Isaac Sim and experiments on a real robot platform.
The evaluation focuses on three aspects:
(i) the accuracy of the learning-based caster-state \emph{estimation and short-horizon prediction}, including predictive uncertainty;
(ii) whether the \emph{horizon-wise weight scheduling of CU-MPC} is activated as intended under increased predictive uncertainty and contributes to reducing caster \emph{misalignment error}; and
(iii) the extent to which the low-level \emph{RL-based execution layer} improves \emph{velocity tracking} and \emph{command executability} under caster-induced contact nonlinearities and external disturbances.






\subsection{Experimental Setup}
\label{subsec:exp_setup}

\subsubsection{Simulation Environment (Isaac Sim)}
We build a differential-drive robot model with freely swiveling casters in Isaac Sim that matches the real platform configuration.
To emphasize caster-induced effects, we design evaluation trajectories that include torque-sensitive maneuvers such as low-speed turning, in-place rotations, and high-curvature cornering.
To reduce potential bias toward a specific path geometry, we conduct the same evaluation over three distinct trajectory scenarios (M/S/C).
In simulation, ground-truth caster angles and caster-related velocities (e.g., swivel/rolling states) are directly available; these signals are used as privileged information for training and as quantitative references for evaluation.


\subsubsection{Baselines (Controllers)}
\label{subsubsec:exp_baselines}
We define the baselines as combinations of a high-level planner (MPC) and a low-level execution layer as follows:
\begin{itemize}
  \item \textbf{B1:} CA-MPC + kinematics controller (PI).
  \item \textbf{B2:} CU-MPC + concurrent-learning execution layer.
  \item \textbf{Proposed:} CU-MPC + the proposed RL execution layer.
\end{itemize}
Here, CA-MPC denotes the baseline caster-aware MPC, and CU-MPC denotes our uncertainty-aware MPC extension (including predictive-uncertainty-driven, horizon-wise weight scheduling).
Baseline B2 replaces only the low-level execution layer with a concurrently trained policy, enabling a controlled comparison with the proposed method under the \emph{same reward function}.

\subsubsection{Implementation and Parameters}
\label{subsubsec:exp_params}
Key parameters used in the learning module (execution layer) and MPC (CA-MPC/CU-MPC)—including horizons, cost weights, and command-mapping settings—are summarized in Table~\ref{tab:exp_params}.





\subsection{Caster-State Estimation and Uncertainty-Aware Scheduling}
\label{subsec:exp_est_pred}

\subsubsection{Caster Angle Estimation Performance}
\begin{figure}
 \vspace{1.0em}
  \centering
  \includegraphics[width=\linewidth]{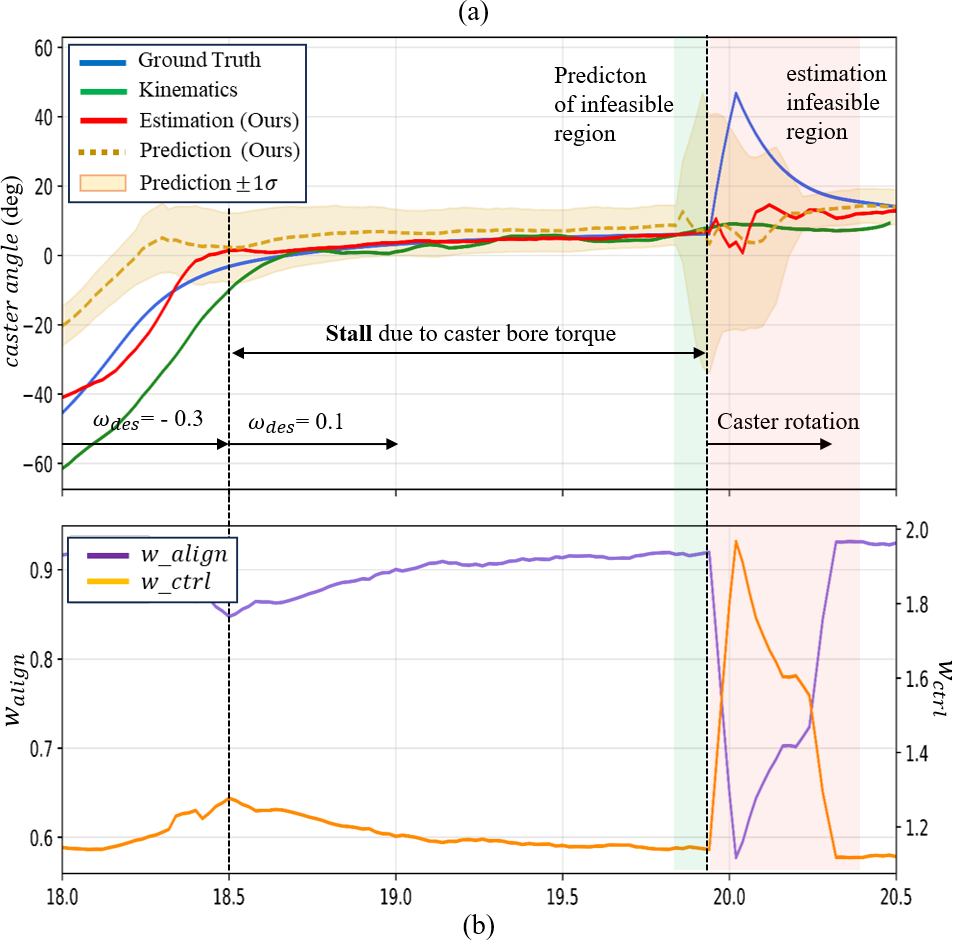}
  \caption{\textbf{Caster estimation/prediction and uncertainty-aware scheduling (in-place low-speed rotation).}
  (a) Left caster angle comparison: GT, kinematic estimator, learned estimator, and learned predictor with the prediction uncertainty band.
  (b) CU-MPC weight scheduling driven by the predicted risk/uncertainty $\rho$: caster-aware weight $w_{\mathrm{cw}}$ and control regularization weight $w_{\mathrm{ctrl}}$.}
  \label{fig:cu_mpc_effect}
\end{figure}

Fig.~\ref{fig:cu_mpc_effect}(a) compares caster angle estimates against ground truth (GT) for the kinematics-based estimator (B1) and the proposed learning-based estimator.
As reported in Table~\ref{tab:sim_results_wide}, the learning-based estimator consistently reduces RMSE$[e_{\hat{\varphi}}]$ across all three path scenarios (M/S/C)
(M: $0.301\!\rightarrow\!0.029$, S: $0.289\!\rightarrow\!0.065$, C: $0.390\!\rightarrow\!0.064$).

This improvement primarily stems from differences in the estimation structure.
The kinematics-based estimator relies on $(v,\omega)$ derived from odometry; in practical systems, odometry is often obtained via EKF fusion of wheel encoders and an IMU, which inevitably introduces filtering delay and can amplify estimation errors during abrupt maneuvers or rapid curvature changes.
In contrast, the proposed method estimates caster states directly from observation history and leverages short-horizon prediction, yielding estimates that are less sensitive to delay and closer to GT.

We note that during rapid caster swing transitions---when the caster overcomes bore torque (e.g., around $t\!\approx\!19.9$\,s in Fig.~\ref{fig:cu_mpc_effect}(a))---both estimators exhibit transient error increases, highlighting the inherent difficulty of accurate estimation under abrupt contact-mode changes and strongly nonlinear regimes.


\begin{figure*}[t]
  \centering
  \includegraphics[width=\linewidth]{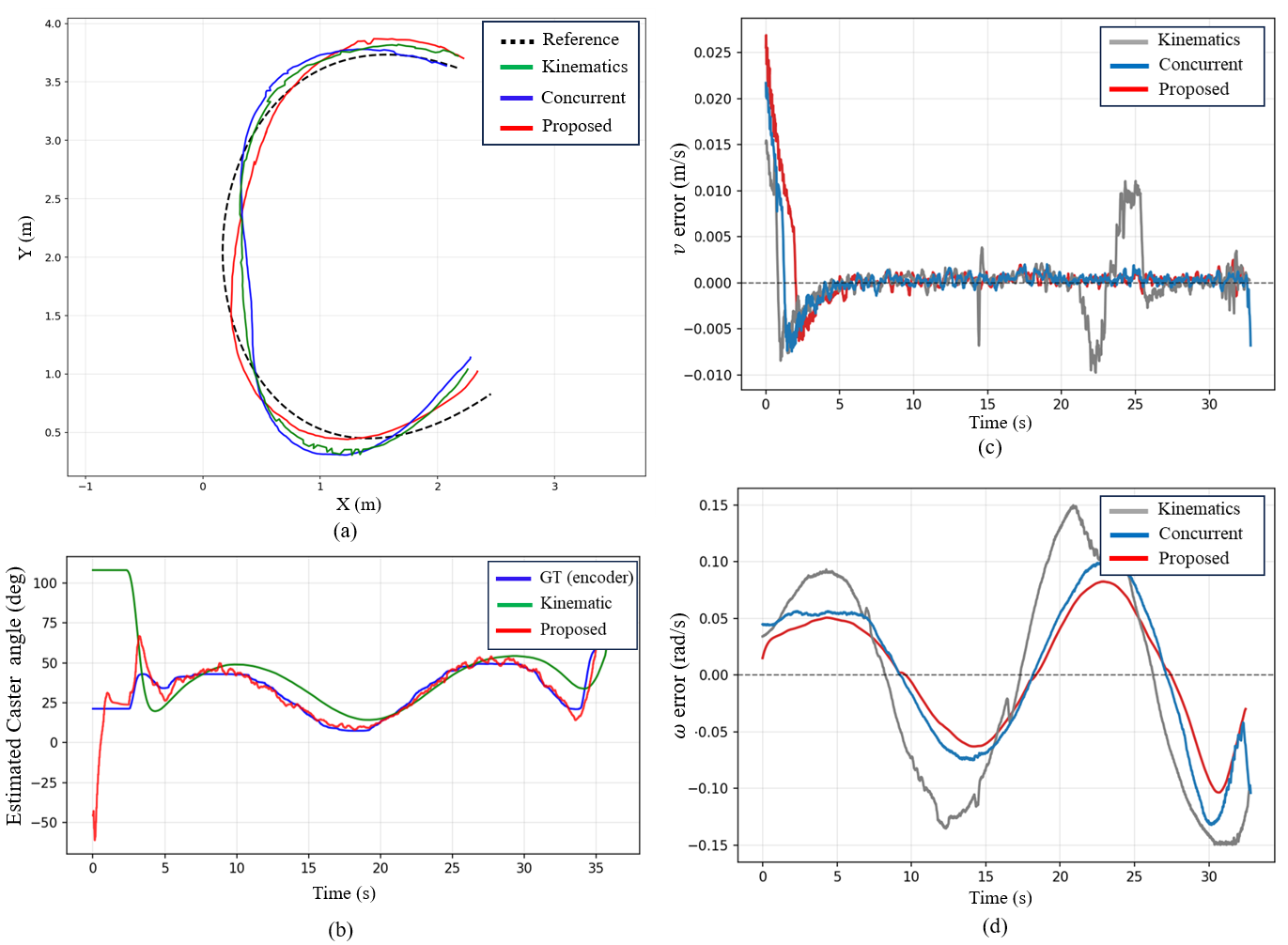}
  \caption{\textbf{Real-world experimental results (C-path).}
  (a) Executed trajectories for the compared methods in the mapped environment.
  (b) Caster angle estimation compared against encoder ground truth (GT), highlighting the phase lag of the kinematics-based estimator and the improved agreement of the proposed learning-based estimator.
  (c) Linear velocity tracking error $e_v$ over time.
  (d) Angular velocity tracking error $e_\omega$ over time.}
  \label{fig:real_experiment_result}
  \vspace{-1.0em}
\end{figure*}

\begin{table}[t]
\centering
\caption{Simulation results in Isaac Sim (M/S/C paths).}
\label{tab:sim_results_wide}
\setlength{\tabcolsep}{2.5pt}
\renewcommand{\arraystretch}{1.12}
\resizebox{\columnwidth}{!}{%
\begin{tabular}{lccccccccc}
\hline
\textbf{Metric} 
& \multicolumn{3}{c}{\textbf{M-path}} 
& \multicolumn{3}{c}{\textbf{S-path}} 
& \multicolumn{3}{c}{\textbf{C-path}} \\
\cline{2-10}
& \textbf{B1} & \textbf{B2} & \textbf{Ours}
& \textbf{B1} & \textbf{B2} & \textbf{Ours}
& \textbf{B1} & \textbf{B2} & \textbf{Ours} \\
\hline
RMSE$[d]$ (m)                         
& 0.154 & \textbf{0.080} & 0.094 & 0.213 & 0.211 & \textbf{0.092} & 0.213 & 0.058 & \textbf{0.043} \\
RMSE$[e_\psi]$ (rad)                  
& 0.532 & \textbf{0.472} & 0.533 & \textbf{0.173} & 0.474 & 0.176 & \textbf{0.127} & 0.184 & 0.129 \\
RMSE$[v]$ (m/s)                       
& \textbf{0.007} & 0.017 & 0.019 & \textbf{0.006} & 0.020 & 0.013 & \textbf{0.006} & 0.013 & 0.011 \\
RMSE$[\omega]$ (rad/s)                
& 0.069 & 0.045 &\textbf{0.029} & 0.074 & 0.057 & \textbf{0.032} & 0.068 & 0.027 & \textbf{0.024} \\
RMSE$[e_{\hat{\varphi}}]$ (rad)          
& 0.301 & 0.276 & \textbf{0.029} & 0.289 & 0.425 & \textbf{0.065} & 0.390 & 0.712 & {0.064}\\
RMSE$[e_{\mathrm{mis}}]$ (rad)
& 0.209 & 0.199 & \textbf{0.108} & 0.208 & 0.250 & \textbf{0.101} & 0.246 & 0.186 & \textbf{0.106} \\
Mean torque (Nm)                      
& \textbf{9.334} & 14.34 & 18.43 & \textbf{10.79} & 13.61 & 15.77 & \textbf{10.58} & 14.36 & 18.56 \\
\hline
\end{tabular}%
}

\vspace{0.25em}
\footnotesize
\noindent
\begin{tabular}{@{}l l@{}}
\multicolumn{2}{@{}l@{}}{\textbf{Metrics (RMSE):}} \\
$d$ & lateral distance-to-path \\
$e_\psi$ & heading error w.r.t.\ the path tangent (rad) \\
$v$, $\omega$ & linear/angular velocity tracking errors \\
$e_{\hat{\varphi}}$ & caster angle estimation error, $e_{\hat{\varphi}}=\hat{\varphi}-\varphi^{\mathrm{GT}}$ (rad) \\
$e_{\mathrm{mis}}$ & caster misalignment error, $e_{\mathrm{mis}}=\varphi-\varphi_{\mathrm{ideal}}(v,\omega)$ (rad)
\end{tabular}
  \vspace{-1.0em} 
\end{table}

\begin{figure}
  \centering
  \includegraphics[width=\linewidth]{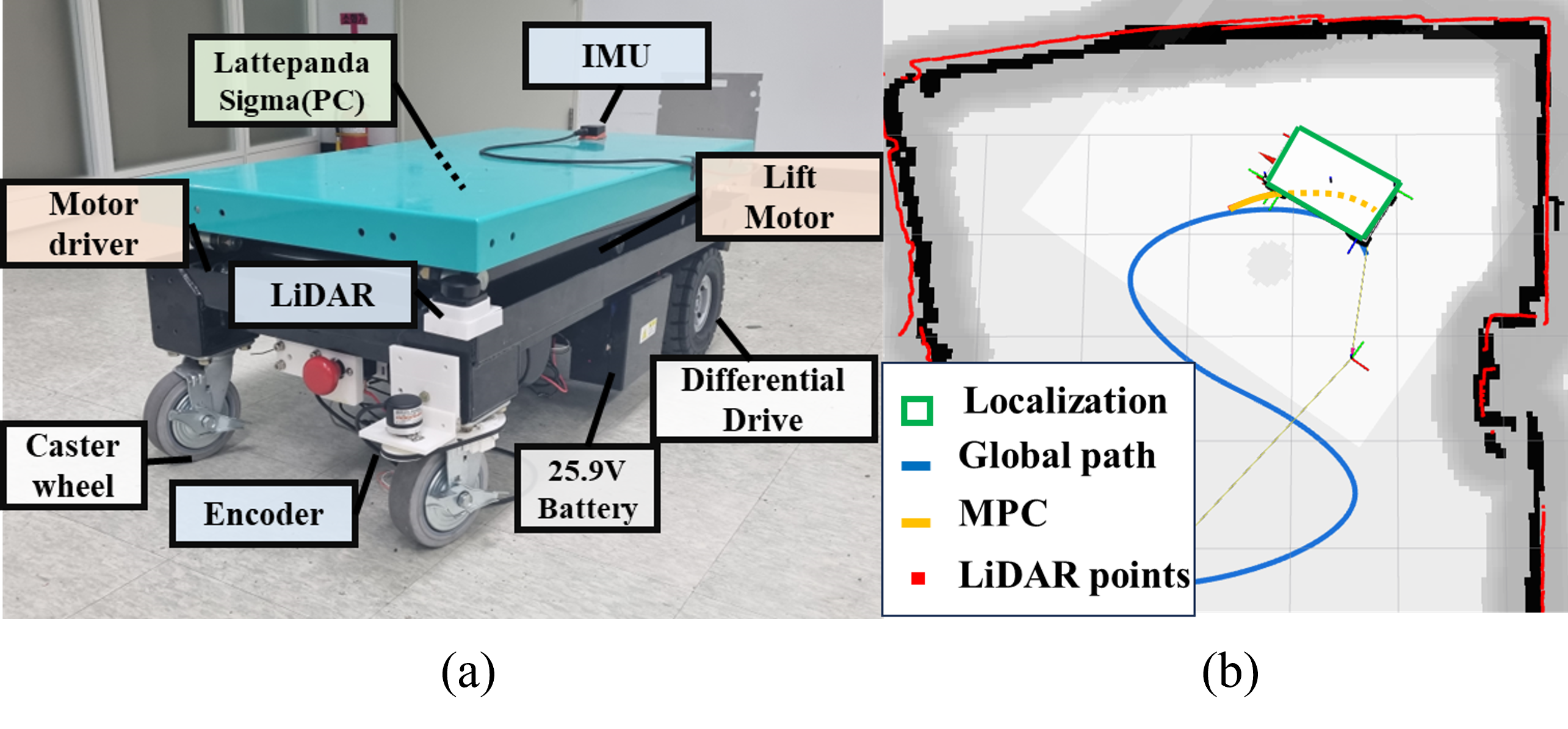}
  \caption{\textbf{Real-world experimental setup.}
  (a) The AMR hardware platform used in the experiments.
  (b) ROS~2-based navigation stack visualization during execution (planning/localization and trajectory tracking).}
  \label{fig:real_setup}
    \vspace{-1.0em} 
\end{figure}
\vspace{0.25em}

\subsubsection{Uncertainty-Aware Scheduling Results}
Fig.~\ref{fig:cu_mpc_effect}(b) shows the predicted risk signal $\rho$ during low-speed in-place rotation and the resulting CU-MPC weight scheduling for $w_{\mathrm{cw}}$ and $w_{\mathrm{ctrl}}$.
Notably, during the rapid caster-swing transition in Fig.~\ref{fig:cu_mpc_effect}(a) (around $t\!\approx\!19.9$\,s), the predictive uncertainty increases; correspondingly, $w_{\mathrm{cw}}$ decreases while $w_{\mathrm{ctrl}}$ increases.
This behavior matches the design intent: when estimation reliability is low, the influence of the caster-aware term is attenuated while input regularization is strengthened, thereby avoiding overly aggressive torque-sensitive maneuvers.

\vspace{0.25em}
\noindent
Quantitatively, Table~\ref{tab:sim_results_wide} shows that the proposed method achieves the lowest misalignment RMSE$[e_{\mathrm{mis}}]$ across all M/S/C paths, supporting that uncertainty-informed scheduling contributes to suppressing caster misalignment.



\subsection{Learning-Based Velocity Tracking Performance}
\label{subsec:exp_tracking}

In this part, we evaluate whether the low-level execution layer can transform the velocity commands generated by the high-level CU-MPC into \emph{physically executable wheel-speed commands} under caster-induced contact effects, thereby stably reproducing the intended $(v,\omega)$.
The tracking performance is quantified by RMSE$[v]$ and RMSE$[\omega]$ in Table~\ref{tab:sim_results_wide}.

\vspace{0.25em}
\noindent
Caster-induced bore torque and contact nonlinearities are major factors that hinder angular-velocity tracking during rotational maneuvers of differential-drive robots.
As shown in Table~\ref{tab:sim_results_wide}, the proposed method achieves the lowest RMSE$[\omega]$ across the M/S/C paths, with values of $0.029/0.032/0.024$, indicating more accurate execution of the high-level angular-velocity commands.
In contrast, RMSE$[v]$ is lower for B1 in some scenarios, suggesting that in our experimental setting the improvement in $\omega$ tracking does not necessarily translate into an equal level of improvement in linear-velocity tracking in the same segments.
Nevertheless, improved $\omega$ tracking contributes to more consistent realization of CU-MPC's velocity intent along paths with frequent turning and misalignment.
Moreover, compared with the concurrent-learning-based execution layer (B2) trained with the same reward function, the proposed method yields lower $\omega$ tracking errors in turning regimes, supporting the effectiveness of the proposed learning scheme.

\noindent
Meanwhile, the mean drive-motor torque tends to increase with the proposed method, which can be interpreted as increased \emph{actuation effort} required to more faithfully realize the target velocity commands under caster-induced effects.
Notably, this increase in effort accompanies a reduction in the misalignment RMSE$[e_{\mathrm{mis}}]$ in Table~\ref{tab:sim_results_wide}, promoting more stable and consistent execution in misalignment-sensitive regimes.
Moreover, the proposed method shows a consistent tendency to reduce the path error RMSE$[d]$ across all M/S/C paths, suggesting that improved reproducibility of velocity commands can also translate into better path-tracking performance.

\begin{table}
\vspace{1.0em} 
\centering
\caption{Real-robot experiment results (C-path).}
\label{tab:real_results_cpath}
\setlength{\tabcolsep}{4pt}
\renewcommand{\arraystretch}{1.12}
\footnotesize
\begin{tabular}{lccc}
\hline
\textbf{Metric} & \textbf{B1} & \textbf{B2} & \textbf{Ours} \\
\hline
RMSE$[d]$ (m)                  & 0.1990 & 0.1171 & \textbf{0.1163} \\
RMSE$[e_\psi]$ (rad)           & 0.3695 & \textbf{0.2113} & 0.2195 \\
RMSE$[v]$ (m/s)                & 0.0083 & 0.0097 & \textbf{0.0059} \\
RMSE$[\omega]$ (rad/s)         & 0.0902 & 0.0719 & \textbf{0.0645} \\
RMSE$[e_{\hat{\varphi}}]$ (rad)   & 0.3450 & 0.2174 & \textbf{0.1005} \\
RMSE$[e_{\mathrm{mis}}]$ (rad) & 0.1880 & 0.1006 &  \textbf{0.0849} \\
Mean current (A)               & 0.7984 &  \textbf{0.7527} & 0.9731 \\
\hline
\end{tabular}
  \vspace{-1.0em} 
\end{table}





\subsection{Real-Robot Validation}
\label{subsec:real_exp}

We validate the feasibility and performance of MSC on a real differential-drive AMR using the same ROS~2 Humble/Nav2-based pipeline as in simulation.
Caster angles are measured as ground truth (GT) using encoders (Fig.~\ref{fig:real_setup}(a)).
A 2D map is built offline with a 2D LiDAR (SLAMTEC S3) and Cartographer, and localization is performed by refining EKF-fused odometry (wheel encoders + IMU, MTi-630R) with AMCL.

\vspace{0.25em}
\noindent
Table~\ref{tab:real_results_cpath} and Fig.~\ref{fig:real_experiment_result} summarize the C-path results.
The proposed method achieves the lowest angular-velocity tracking error (RMSE$[\omega]=0.0645$) and the lowest path error (RMSE$[d]=0.1163$), indicating improved realization of the high-level velocity intent in practice.
In Fig.~\ref{fig:real_experiment_result}(b), the kinematics-based estimator exhibits a clear phase lag, whereas the proposed learning-based estimator attains the best agreement with encoder GT (RMSE$[e_{\hat{\varphi}}]=0.1005$).
The mean current increases to $0.9731$\,A, consistent with higher actuation effort required to track the commanded motion under real contact effects.





\section{Conclusion}
\label{sec:conclusion}

This paper proposed MSC, a planning--execution integrated hierarchical control framework that mitigates the caster-induced planning--execution gap in low-speed turning and high-curvature maneuvers.
MSC integrates (i) an RL-based execution layer for caster-state estimation/prediction with uncertainty, (ii) a caster-uncertainty-aware NMPC (CU-MPC) with horizon-wise risk-based weight scheduling, and (iii) a policy that transforms high-level $(v,\omega)$ commands into executable wheel-speed commands.

Across simulation paths, MSC reduced caster-angle estimation and misalignment errors and improved angular-velocity tracking, with increased actuation effort.
Real-robot experiments showed consistent gains in caster-angle estimation and in tracking/path errors compared to baselines.

\indent
In summary, the caster-induced planning--execution gap can be mitigated more effectively when \emph{uncertainty-aware planning} at the high level and \emph{contact-consistent command realization} at the execution level are considered jointly.
While addressing caster effects at a single layer can be beneficial, our results suggest that a hierarchical design that consistently accounts for caster states (and their uncertainty) provides more reliable improvements across operating regimes.
Moreover, leveraging learning-based estimation/prediction with uncertainty quantification under partial observability proved effective in both simulation and real-robot experiments, indicating that the proposed learning-based approach is a practical and impactful component for this problem setting.


\end{document}